\pdfoutput=1

\documentclass[11pt]{article}

\usepackage{rotating}
\usepackage{tablefootnote}
\usepackage[most]{tcolorbox}
\usepackage{enumitem}
\usepackage{xcolor}
\usepackage{threeparttable}
\usepackage{url}
\usepackage[final]{acl}
\usepackage{tcolorbox}
\usepackage{enumitem}
\usepackage{times}
\usepackage{latexsym}
\usepackage{tabularx}
\usepackage{url}
\usepackage{amsfonts}
\usepackage{amsmath}
\usepackage{rotating}
\usepackage{booktabs} 
\usepackage{colortbl}
\usepackage{multirow} 
\usepackage{siunitx}  
\usepackage{caption}  
\usepackage[T1]{fontenc}
\usepackage{booktabs}
\usepackage[utf8]{inputenc}
\usepackage{mdframed}
\usepackage{xcolor}

\mdfdefinestyle{examplebox}{
  backgroundcolor=gray!10,
  linecolor=black!50,
  linewidth=0.8pt,
  roundcorner=2pt,
  innerleftmargin=4pt,
  innerrightmargin=4pt,
  innertopmargin=4pt,
  innerbottommargin=4pt
}
\usepackage{microtype}

\usepackage{pgfplots}
\pgfplotsset{compat=1.18}
\usepackage{pgfplotstable}
\usepackage{filecontents}
\usepackage{adjustbox}
\usetikzlibrary{patterns, positioning}
\usepackage{booktabs}

\usepackage{inconsolata}

\usepackage{graphicx}

\usepackage{amssymb}

\usepackage{cleveref}
\title{Afrispeech Semantics: Evaluating Audio–Semantic Reasoning in Spoken Language Models Across Domains and Accents}

\author{
  Chibuzor Okocha \\
  University of Florida \\
  \texttt{c.okocha@ufl.edu} \\
  \And
  Christan Grant \\
  University of Florida \\
  \texttt{christan@ufl.edu}
}

\begin{document}
\maketitle
\begin{abstract}
Audio language models (ALMs) are increasingly used for speech-based understanding; yet, their ability to perform semantic reasoning beyond transcription, Text-to-Audio Retrieval, Captioning, and Question-Answering accuracy remains insufficiently benchmarked. In particular, the effects of accent variation, domain shift, and semantic over-inference on audio reasoning are poorly understood.  We evaluate audio language models across five semantic and paralinguistic reasoning tasks: entailment, consistency, plausibility, accent drift, and accent restraint. Collectively, these tasks assess a model’s ability to reason over spoken audio as the primary evidence source, including whether a textual hypothesis can be inferred, contradicted, or left undetermined by the audio, whether statements align or conflict with spoken content, whether claims are plausible given the discourse, and whether model predictions remain stable or appropriately constrained across accent variation. These findings highlight critical limitations in current audio reasoning evaluations and hope to provide guidance for more robust and equitable ALM design and assessment.

\end{abstract}

\section{Introduction}
Recent multimodal models, commonly referred to as audio language models (ALMs), are trained on large collections of audio–text pairs using either contrastive learning  \cite{elizalde_clap_2023} or next-token prediction objectives \cite{qwen2, tang2023salmonn, goel2025, kimiteam2025, yusalmonn, ghosh_gama_2024}. Once trained, ALMs can be prompted to perform a wide range of tasks grounded in audio, including captioning, retrieval, and question answering, and have demonstrated strong performance across many established benchmarks \cite{sakshi2024, wang2024audiobench, openai2024}.

Despite these advances, most existing evaluations primarily focus on surface-level correctness rather than semantic reasoning grounded in the audio signal \cite{wang2024audiobench, yang2025b, yang2024b, gao2025a}. In open-ended settings, ALMs are often rewarded for producing plausible responses, even when those responses rely on contextual assumptions or linguistic priors rather than evidence present in the audio \cite{chiang2025}. This limitation becomes particularly problematic for interactive and reasoning-oriented applications, where models are expected to infer what can—and cannot—be concluded from what is heard \cite{sanni_afrispeech-dialog_2025}.

To address this gap, prior work introduced audio entailment as a focused task for evaluating deductive reasoning in audio language models \cite{deshmukh2023}, framing the problem as determining whether a textual hypothesis is entailed, contradicted, or unsupported by an audio premise. While this formulation provides a principled starting point, it captures only a narrow slice of the reasoning challenges faced by ALMs. Existing benchmarks are limited in domain diversity and do not explicitly test failure modes such as semantic over-inference, robustness to unfamiliar named entities, or sensitivity to accent and pronunciation variation \cite{shi2024, wang2024audiobench, sakshi2024}.

In this work, we expand the evaluation of audio reasoning beyond a single entailment task or deductive reasoning task. We introduce a unified semantic reasoning framework comprising several domain-diverse speech datasets, each paired with task formulations tailored to the semantic properties of the domain.

Using this framework, we benchmark state-of-the-art contrastive and next-token prediction audio language models under a controlled inference protocol. Our results reveal consistent patterns of over-entailment, domain-specific reasoning failures, and accent-conditioned semantic drift, even when transcription quality is high. These findings suggest that current benchmarks substantially underestimate reasoning errors in audio language models, underscoring the need for more comprehensive, domain-aware evaluation of semantic reasoning from audio.

In this work, we study logical and semantic reasoning for ALMs. Our contributions are as follows:

\begin{itemize}

    \item We propose multiple task formulations designed to test whether models preserve meaning under pronunciation variation rather than relying on contextual priors.

    \item Hypotheses are generated using LLaMA model and systematically verified and corrected by human annotators, ensuring semantic validity and grounding in the audio evidence, including accented and unfamiliar speech patterns.

    \item We benchmark both contrastive and next-token prediction audio language models under a unified inference protocol, revealing consistent over-entailment and accent-sensitive reasoning failures that are not captured by existing benchmarks.
\end{itemize}

\section{Related works}
Audio–language models combine acoustic perception with language modeling to support a broad range of audio understanding tasks \cite{ghosh_gama_2024, deshmukh2023}. Early approaches focused primarily on contrastive learning to align audio and text embeddings, enabling strong performance on retrieval and classification tasks \cite{elizalde_clap_2023}. More recent work has explored next-token prediction frameworks that treat audio understanding as a conditional text-generation problem, enabling models to perform open-ended tasks such as captioning, dialogue, and question answering \cite{deshmukh2024a, deshmukh, deshmukh2023}. These advances have substantially expanded the functional scope of ALMs; however, evaluation has largely emphasized task completion or linguistic plausibility rather than the semantic validity of the inferred conclusions \cite{peng2025}.

As ALMs have become more capable of generating free-form responses, concerns have emerged regarding their reliance on contextual priors and language statistics rather than evidence grounded in the audio signal \cite{sakshi2024, yang2025a}. Existing audio question answering and captioning benchmarks often permit multiple acceptable outputs, making it difficult to distinguish between correct inference and plausible hallucination \cite{chu2023}. As a result, models may appear to perform well despite systematically over-interpreting or misattributing audio events \cite{shi2024, kubis2025}. These limitations have motivated interest in more principled evaluation frameworks that explicitly test reasoning behavior rather than surface-level alignment \cite{kubis2025}.

Audio entailment was introduced as a structured task to evaluate deductive reasoning in audio–language models by determining whether a textual hypothesis is entailed, contradicted, or unsupported by an audio premise \cite{deshmukh}. By framing audio understanding as a three-way inference problem, this work provided an important step toward disentangling reasoning from generation quality and revealed significant reasoning deficiencies in state-of-the-art models. However, the focus of audio entailment is intentionally narrow: it evaluates a single reasoning formulation over a limited set of domains and does not explicitly probe how reasoning failures vary across domain types, semantic phenomena, or speech characteristics such as accent variation.

Related work in vision–language and text-based inference has shown that single-task benchmarks can underestimate reasoning errors and mask systematic failure modes \cite{sadasivan2025}. Multi-task and domain-aware evaluation has proven essential for uncovering issues such as over-generalization, reliance on world knowledge, and sensitivity to spurious correlations \cite{afrispeech, sanni_afrispeech-dialog_2025}. In the audio domain, however, comparable multitask semantic-reasoning evaluations remain underdeveloped. Prior benchmarks do not systematically test semantic restraint, robustness to unfamiliar named entities, or accent-conditioned meaning drift phenomena that are especially relevant for spoken language understanding in diverse, real-world settings \cite{wang2025a}.

\begin{table}[t]
\centering
\small
\begin{tabular}{lll}
\hline
\textbf{Task} & \textbf{Reasoning} & \textbf{Complexity}  \\
\hline
Entailment & Deductive semantic & $\blacksquare\blacksquare\blacksquare\blacksquare$ \\
Consistency & Logical compatibility & $\blacksquare\blacksquare$ \\
Plausibility & Pragmatic semantic & $\blacksquare\blacksquare$  \\
Accent Drift & Robustness &$\blacksquare\blacksquare\blacksquare$ \\
Accent Restraint & Causal restraint & $\blacksquare\blacksquare\blacksquare\blacksquare\blacksquare$ \\
\hline
\end{tabular}
\caption{Task taxonomy by reasoning type and the complexity of each reasoning type.}
\label{tab:reasoning-taxonomy}
\end{table}

\begin{table}[t]
\centering
\small
\begin{tabular}{lcc}
\toprule
\textbf{Dataset} & \textbf{Hours (h)} & \textbf{Accent Coverage} \\
\midrule
AfriSpeech-200\tablefootnote{\url{https://huggingface.co/datasets/intronhealth/afrispeech-200}} & 200 & 13 \\
AfriSpeech-General\tablefootnote{\url{https://huggingface.co/datasets/intronhealth/afrispeech-dialog}} & 2.07 & 8 \\
Afri-Names\tablefootnote{\url{https://huggingface.co/datasets/intronhealth/afri-names}} & 8.92 & 12  \\
Afrispeech - Medical\tablefootnote{\url{https://huggingface.co/datasets/intronhealth/med-convo-nig}} & 4.20 & 6 \\
\bottomrule
\end{tabular}
\caption{Overview of the four speech corpora used in this work.  Durations are measured after pruning unusable segments. Detailed statistics and split methodology are provided in Appendix~\ref{sec:appendix_datasets}.}
\label{tab:datasets-main}
\end{table}

\section{Semantic Reasoning Tasks for Audio Language Models}
\label{sec:tasks}

\subsection{Problem Setup and Notation}

We study semantic reasoning in audio language models through tasks that require models to draw conclusions from spoken audio. Each task is formulated as a relationship between an audio recording and a textual hypothesis. Let $a$ denote an audio recording serving as the \emph{premise}, and let $h$ denote a natural language \emph{hypothesis}. Given a pair $(a, h)$, the model predicts the semantic relationship between the hypothesis and the content expressed in the audio.
An overview of the reasoning types associated with each task is provided in Table~\ref{tab:reasoning-taxonomy}.

Following prior work on audio entailment \cite{deshmukh}, we restrict inference to information supported by the audio signal itself. Models are not permitted to assume unstated facts or rely on external world knowledge beyond what can be reasonably inferred from the spoken content.

\subsection{Audio Entailment}

We adopt audio entailment as the core semantic reasoning task. The objective is to determine whether a hypothesis is supported, contradicted, or left undetermined by an audio premise. Specifically, each $(a, h)$ pair is assigned one of three labels:
\begin{itemize}
    \item \textbf{Entailment (E):} The audio provides sufficient evidence to support the hypothesis.
    \item \textbf{Neutral (N):} The audio does not provide enough information to determine the truth of the hypothesis.
    \item \textbf{Contradiction (C):} The audio provides sufficient evidence to refute the hypothesis.
\end{itemize}

Formally, the task is defined as
\[
f(a, h) \rightarrow y, \quad y \in \{E, N, C\}.
\]

This task evaluates deductive semantic reasoning grounded in spoken language. Unlike text-based entailment, audio entailment requires models to reason over acoustic realizations of meaning, including speaker variation and prosodic cues, while maintaining strict evidence-based inference.

\subsection{Plausibility and Consistency}

Beyond entailment, models may conflate semantic compatibility with evidential support. We therefore introduce plausibility and consistency tasks that further probe the boundaries of inference.

In the \textbf{plausibility} task, hypotheses are constructed to be reasonable given commonsense knowledge or discourse context, but are neither stated nor implied by the audio. The task assesses whether models incorrectly accept such hypotheses based solely on plausibility.

In the \textbf{consistency} task, hypotheses are either semantically compatible or incompatible with the spoken content. Unlike entailment, this task does not admit a neutral option; instead, it focuses on detecting agreement or contradiction with the audio premise. Together, these tasks assess whether models rely on semantic evidence rather than commonsense priors when making judgments.

\subsection{Accent-Conditioned Semantic Drift and Accent Restraint}

Spoken language varies substantially across speakers and accents. To evaluate robustness under such variation, we introduce tasks that examine accent-conditioned semantic drift. In this setting, audio recordings differ in accent or pronunciation while preserving equivalent semantic content, and hypotheses remain unchanged. The task assesses whether accent variation systematically alters semantic predictions.

We further introduce an accent restraint task to test whether models appropriately suppress accent-based cues when the accent is semantically irrelevant. Here, accent functions as a nuisance variable, and correct behavior requires invariant semantic judgments across accented realizations.

\section{Dataset Construction and Annotation}
\label{sec:dataset}

\begin{figure*}[t]
  \centering
  \includegraphics[width=\linewidth]{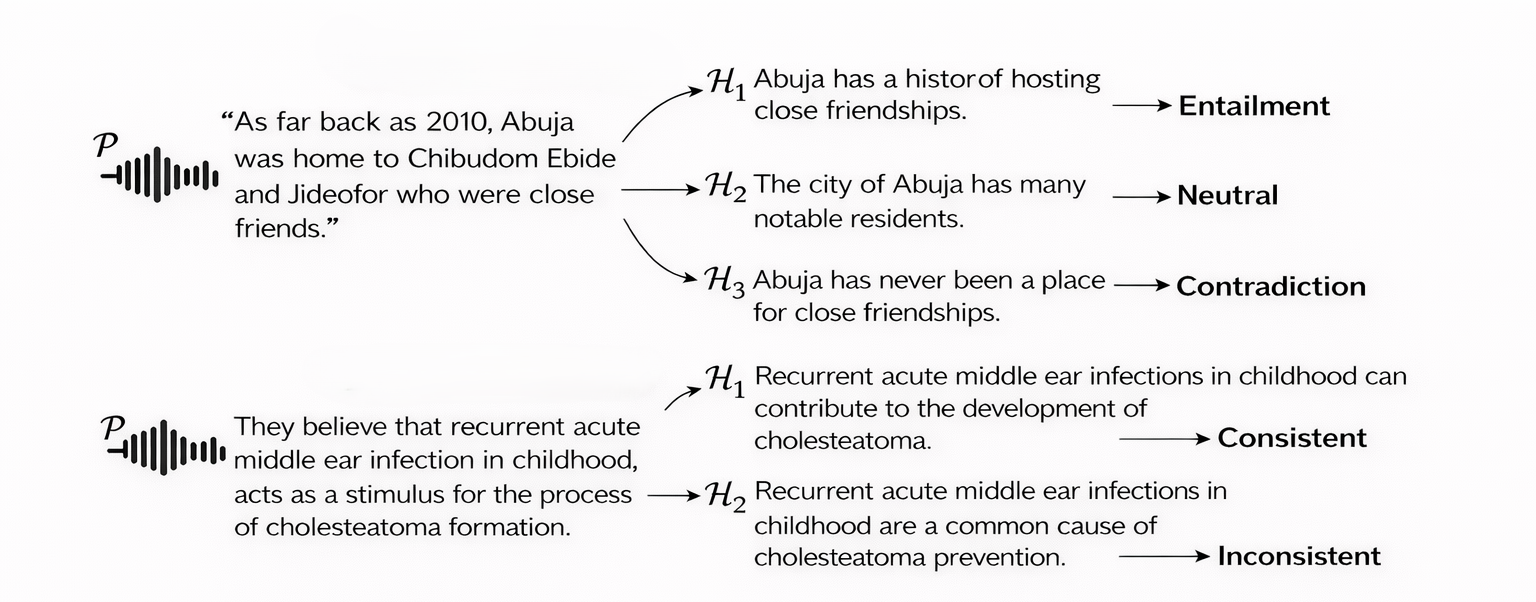}
  \caption{Examples of two AfriSpeech-derived audio reasoning tasks. Each example uses an audio premise $P$ (shown as a waveform icon and the verbatim transcript) paired with hypotheses $H_1, H_2,$ and $H_3$ (or fewer, depending on the task). For Spoken Entailment, hypotheses are labeled \textsc{Entailment}, \textsc{Neutral}, or \textsc{Contradiction}. For Medical Consistency, hypotheses are labeled \textsc{Consistent} or \textsc{Inconsistent}.}
  \label{fig:afrispeech-task-examples}
\end{figure*}

In this section we describe how we curated and annotated the data used throughout our evaluation.  Rather than treating this step as a minor implementation detail, we view careful dataset construction as central to any fair benchmark.  Our goal was to build a suite of tasks that capture real-world variation in both content and pronunciation, while ensuring that every hypothesis is grounded in the acoustic evidence.  Table~\ref{tab:datasets-main} gives a high-level overview of the four underlying corpora.  A more detailed description of the speaker demographics, transcript quality and split methodology can be found in appendix~\ref{sec:appendix_datasets}.  We provide qualitative examples of these tasks in Figure~\ref{fig:afrispeech-task-examples}.

\subsection{Audio Premises}
Each dataset is built around a collection of audio premises $a$ paired with one or more textual hypotheses.  The corpora span distinct domains, including conversational telephone speech, read speech focused on named entities, and clinical dialogues to elicit a variety of reasoning behaviors.  Importantly, the recordings cover a wide range of African accents and dialects, meaning models must contend with both linguistic and paralinguistic variation rather than idealised laboratory speech.  We selected datasets whose original annotations were created from scratch by listening to the audio, thereby avoiding the confounding effect of textual metadata.  Further details on duration, speaker demographics, and accent distribution are provided in appendix~\ref{sec:appendix_datasets}.

\subsection{Hypothesis Generation}
For every audio premise, we create a small set of hypotheses that probe different semantic relationships: whether a statement is entailed by the audio, contradicted by it, merely plausible under common sense, or tests for accent-induced drift.  To scale this process, we first use an LLM to propose candidate hypotheses under carefully crafted prompts that forbid unsupported negation and extraneous world knowledge.  These prompts are reproduced verbatim in appendix~\ref{app:prompts}.  Crucially, the LLM proposals are only a starting point.

\subsection{Human Verification and Correction}
After automatic generation, each candidate hypothesis is vetted by trained human annotators.  Annotators listen to the audio in full and judge whether the proposed statement is entailed, contradicted, or unsupported by the recording.  If a hypothesis contains hallucinated details or ambiguous phrasing, annotators edit or replace it to ensure that the final set of hypotheses is both semantically precise and audibly grounded.  This manual verification is particularly critical when dealing with accented speech or rare proper names, where large language models often introduce subtle semantic drift.  Our annotation protocol, including guidelines, quality control procedures, and examples, is described in appendix~\ref{sec:appendix_humanverification}.

\section{Experimental Setup}
\label{sec:experiments}

This section describes the models, inference protocols, and evaluation procedures used to assess semantic reasoning in audio language models. Our setup is designed to enable fair comparison across model families while isolating reasoning behavior from transcription or generation quality.

\subsection{Audio Language Models}
We evaluate two broad classes of audio language models (ALMs): \emph{contrastive} models and \emph{next-token prediction} models. Contrastive ALMs learn joint audio--text representations and are commonly used for retrieval and classification tasks \cite{elizalde_clap_2023}. Next-token prediction ALMs condition on audio and text inputs to generate free-form textual responses \cite{deshmukh2023a}. All models used are open-source and state-of-the-art.

For contrastive models, the audio premise and hypothesis are encoded separately, and their similarity is used to predict the semantic relationship between them. For next-token prediction models, the audio premise and hypothesis are provided as input, and the model generates a textual response indicating its judgment. We evaluate 10 audio language models in total: 3 constractive models and 7 next-token prediction models.

We evaluate all models in a zero-shot setting unless otherwise stated.  Model specifications and checkpoints are summarised in Table~\ref{tab:model_specs}, and inference hyperparameters are listed in Table~\ref{tab:hyperparameters}. These tables provide the necessary context to reproduce our experiments.

\subsection{Prompt Selection}
\label{sec:prompt_selection}
We observed that some models are sensitive to prompt phrasing for the semantic consistency task.
To avoid cherry-picking, we pre-defined three prompt variants and selected the variant that achieved the best macro F1 on a held-out development subset, aggregated across datasets.
Table~\ref{tab:prompt_ablation_af3} reports the comparison, and Appendix~\ref{app:prompts} lists the full prompt templates used.
We use the selected prompt variant for all evaluations.

\subsection{Inference Protocol}
To ensure comparability across model types, we adopt a unified inference protocol aligned with the task formulation in Section~\ref{sec:tasks}. For contrastive models, similarity scores between audio and hypothesis embeddings are mapped to entailment, neutral, or contradiction labels using fixed thresholds derived from a validation set. Threshold selection procedures follow standard practice and are detailed in the \cref{app:prompts}.

\subsection{Evaluation and Label Mapping}
Evaluating free-form outputs from next-token prediction models presents challenges due to variability in phrasing and instruction-following behavior. To address this, we adopt a model-based evaluation strategy in which a lightweight language model maps generated responses to discrete labels. This approach is validated against a human-annotated subset and achieves high agreement.

For contrastive models, predictions are obtained directly from thresholded similarity scores, eliminating the need for additional post-processing. All evaluations are conducted at the hypothesis level and aggregated at the dataset and task level.

\subsection{Metrics}
We report standard classification metrics, including accuracy, precision, recall, and macro-averaged F1 score. For multi-hypothesis settings, metrics are computed across all hypotheses associated with an audio premise. In addition, we report class-wise performance to analyze asymmetric error patterns, such as over-entailment and under-detection of contradiction.

To examine how reasoning performance varies across task types, we report results separately for entailment, semantic restraint, plausibility consistency, and accent-conditioned semantic drift tasks. Detailed breakdowns and statistical summaries are provided in the \cref{app:evaluation_metrics}.

\subsection{Reproducibility}
All datasets, fixed splits, inference prompts, and evaluation scripts will be released to support reproducibility. Hyperparameters and implementation details necessary to replicate our results are documented in the \cref{tab:hyperparameters}.

\section{Results}
\label{sec:results}

To interpret the scores listed in the following tables, we first provide a high-level summary of model behaviour.  We evaluate eleven generative audio language models and three contrastive models across four corpora and five semantic reasoning tasks.  Our primary metrics are accuracy and macro--averaged F1, supplemented by class-specific accuracies to detect asymmetric error patterns.  In the aggregate, generative models substantially outperform contrastive approaches on tasks that require nuanced reasoning, though contrastive baselines offer useful lower bounds.

\begin{figure*}[t]
  \centering
  \small
  \includegraphics[width=\linewidth]{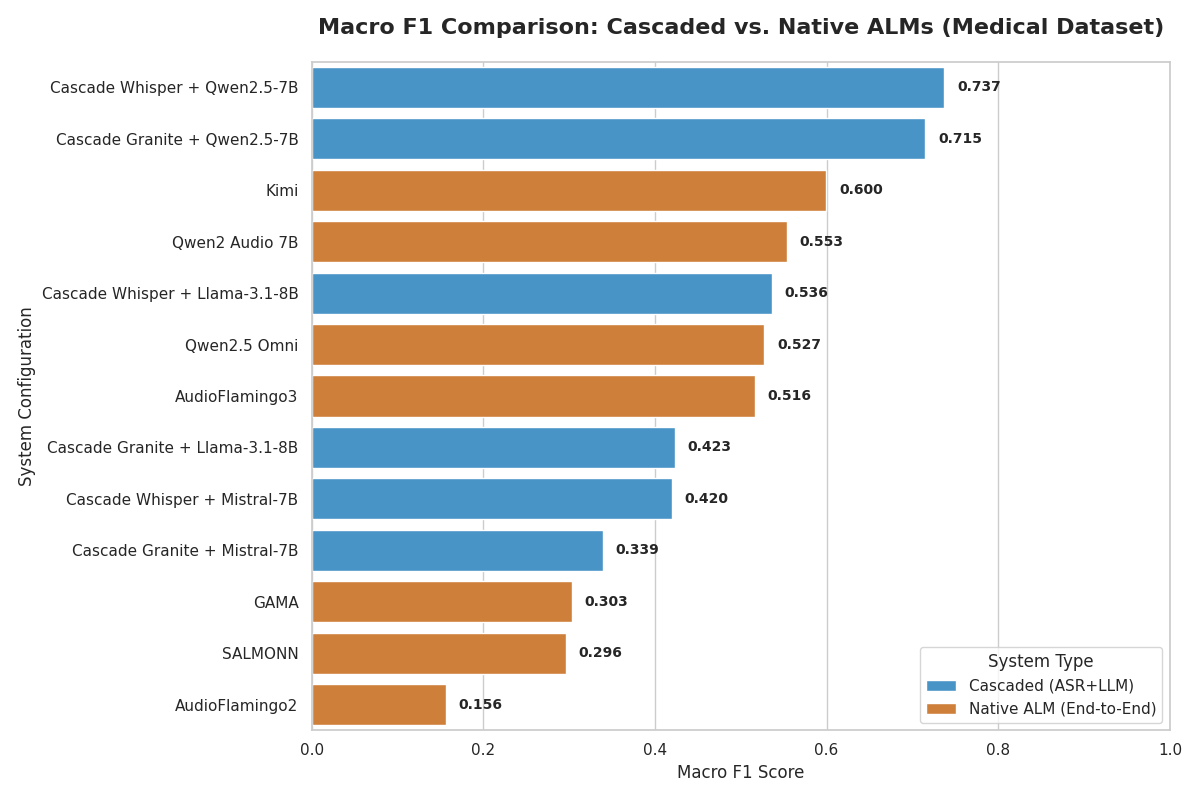}
  \caption{Comparative analysis of Macro F1 scores for Cascaded (ASR+LLM) systems and Native Audio Language Models (ALMs) on the Medical Audio Entailment task.}
  \label{fig:afrispeech-task-examples}
\end{figure*}

\begin{table}[t]
\centering
\small
\begin{tabular*}{\columnwidth}{@{\extracolsep{\fill}} l cccc}
\toprule
\textbf{Cascaded System} & \textbf{Acc} $\uparrow$ & \textbf{F1} $\uparrow$ & \textbf{E-Acc} $\uparrow$ & \textbf{C-Acc} $\uparrow$ \\
\midrule
W + Llama-3.1-8B & 0.560 & 0.536 & 0.913 & 0.189 \\
W + Mistral-7B & 0.469 & 0.420 & 0.333 & 0.927 \\
W + Qwen2.5-7B & 0.739 & \textbf{0.737} & 0.870 & 0.539 \\
\midrule
G + Llama-3.1-8B & 0.478 & 0.423 & 0.942 & 0.088 \\
G + Mistral-7B & 0.425 & 0.339 & 0.145 & 0.986 \\
G + Qwen2.5-7B & 0.715 & 0.715 & 0.710 & 0.625 \\
\bottomrule
\end{tabular*}
\caption{Summarized performance of Cascaded (ASR+LLM) systems on Audio Entailment, where W is for whisper large v3 and G is for IBM Granite speech. Results are averaged across all difficulty levels.}
\label{tab:cascade_nli_summary}
\end{table}

\subsection{Overall performance across tasks}

Across all datasets, we observe a wide spread of performance.  On the AfriSpeech-200 corpus, the strongest generative models (e.g., Qwen2.5Omni and Qwen2AudioInstruct) achieve macro F1 scores above 0.82 on plausibility and consistency tasks, while weaker baselines such as SALMONN lag markedly.  Results on the AfriSpeech-Gen and Medical sets show similar trends, with Qwen2-based models achieving F1 scores around 0.85--0.95 in many settings.  The contrastive models (LAION-CLAP and MSCLAP variants) perform near chance on most reasoning tasks, highlighting the limitations of embedding-only approaches for fine-grained semantics.  Detailed per-task and per-model results are tabulated below.  Qualitative examples comparing model outputs on these tasks are presented in Table~\ref{tab:qualitative_examples} (Appendix~\ref{sec:appendix_qualitative}).

\subsection{Performance of Cascaded Systems}
Our evaluation of cascaded architectures combining state-of-the-art ASR with powerful Large Language Models (LLMs) reveals a significant performance advantage over native end-to-end models for the audio entailment task. As summarized in Table~\ref{tab:cascade_nli_summary}, the Cascade Whisper + Qwen2.5-7B configuration achieved the highest overall performance with a Macro F1 of 0.737, representing the current benchmark for our medical consistency dataset.

Interestingly, we observe a distinct "behavioral split" among different LLM backbones. Systems utilizing Llama-3.1-8B demonstrated high Entailment Accuracy (E-Acc over 0.91), whereas Mistral-7B variants were exceptionally robust at identifying contradictions (C-Acc near 0.93--0.98). This suggests that the choice of LLM in a cascaded pipeline allows for tuning the system toward specific safety or verification goals, such as prioritizing the detection of medical misinformation.

The comparative strengths of these configurations are visualized in Figure~\ref{fig:afrispeech-task-examples}. The chart shows that both Whisper- and Granite-based cascades consistently outperform native ALMs such as Kimi and AudioFlamingo3 on medical data. This performance gap suggests that for complex semantic reasoning, the precision afforded by a discrete transcription step remains superior to the direct audio-to-text latent mapping currently utilized by most end-to-end audio language models.

\subsection{Task-wise analysis}
\paragraph{Audio Plausibility and Consistency.}  The audio plausibility task measures whether models can distinguish statements that are plausible but unsupported by the audio from those that are implausible.  As shown in Tables~\ref{tab:plausibility_generative_models_results} and~\ref{tab:plausibility_constructive_models_results}, generative models tend to accept plausible but unsupported statements, with accuracy ranges from 0.41 to 0.97.  Qwen2.5Omni and AudioFlamingo models often over-accept plausible statements, indicating a propensity for hallucination.  Consistency, on the other hand, probes whether a hypothesis aligns or conflicts with the audio without a neutral option.  Generative models again dominate, with Kimi-Audio and Qwen2.5Omni achieving F1 scores above 0.8 on AfriSpeech-Gen (Table~\ref{tab:consistency_generative_models_results}).
Full results for contrastive baselines on these tasks are provided in Table~\ref{tab:plausibility_constructive_models_results} and Table~\ref{tab:consistency_constructive_models_results}.

\paragraph{Audio Entailment.}  The audio entailment task asks whether a hypothesis is entailed, contradicted, or unsupported by the audio premise.  Table~\ref{tab:nli_generative_models_results} shows that even the best models struggle on this three-way classification: F1 scores rarely exceed 0.71 on AfriSpeech-200, with frequent confusion between neutral and contradiction.  Contrastive models perform poorly, reflecting their limited capacity to discriminate fine-grained semantic relations (Table~\ref{tab:nli_constructive_models_results}).  Qwen2-based models outperform others, but still exhibit a tendency to over-entail, as evidenced by lower neutral and contradiction accuracies.

\paragraph{Accent Drift and Restraint.}  The AfriNames subset isolates reasoning under accent variation and low-semantic-content utterances.  Table~\ref{tab:afrinames_accent_drift} reports bias rates (fraction of accent-sensitive inferences) and accept rates (fraction of accent-invariant predictions).  Qwen2AudioInstruct shows the lowest bias (33.25\%) while AudioFlamingo2 exhibits substantial accent bias (96.75\%).  The restraint evaluation in Table~\ref{tab:afrinames_restraint} underscores that models often hallucinate unsupported content; SALMONN and LAION-CLAP produce hallucinations in more than 60\% of cases.  Models such as Qwen2.5Omni strike a better balance between suppressing hallucination and affirming supported content, achieving a semantic restraint accuracy (SRA) of 66.0\%.

\begin{table*}[t]
\centering
\small
\begin{tabular}{llccccccc}
\toprule
\textbf{Dataset} & \textbf{ALM} & \textbf{Acc} $\uparrow$ & \textbf{P} $\uparrow$ & \textbf{R} $\uparrow$ & \textbf{F1} $\uparrow$ & \textbf{Acc\_P} $\uparrow$ & \textbf{Acc\_I} $\uparrow$ \\
\midrule
\multirow{8}{*}{Afri-200} & AudioFlamingo2 & 0.2585 & 0.2548 & 0.2573 & 0.2560 & 0.5146 & 0.0000 \\
 & AudioFlamingo3 & 0.9512 & 0.9520 & 0.9511 & \textbf{0.9512} & 0.9709 & 0.9314 \\
 & GAMA & 0.4390 & 0.8211 & 0.4374 & 0.4469 & 0.7670 & 0.1078 \\
 & Kimi & 0.6927 & 0.7878 & 0.6941 & 0.6659 & 0.4078 & 0.9804 \\
 & Qwen2.5 Omni & 0.8293 & 0.8723 & 0.8301 & 0.8244 & 0.6602 & 1.0000 \\
 & Qwen2 Audio 7B & 0.8146 & 0.8722 & 0.8155 & 0.8180 & 0.6408 & 0.9902 \\
 & SALMONN & 0.6439 & 0.7630 & 0.6455 & 0.5999 & 0.3107 & 0.9804 \\
\addlinespace[0.5em]
\multirow{8}{*}{Afri-Gen} & AudioFlamingo2 & 0.3707 & 0.2590 & 0.3707 & 0.3050 & 0.7414 & 0.0000 \\
 & AudioFlamingo3 & 0.9138 & 0.9552 & 0.9138 & 0.9339 & 0.9310 & 0.8966 \\
 & GAMA & 0.4741 & 0.8597 & 0.4741 & 0.5529 & 0.7414 & 0.2069 \\
 & Kimi & 0.8103 & 0.9194 & 0.8103 & 0.8506 & 0.9655 & 0.6552 \\
 &Qwen2.5 Omni & 0.9569 & 0.9570 & 0.9569 & \textbf{0.9569} & 0.9483 & 0.9655 \\
 & Qwen2 Audio 7B & 0.8103 & 0.8523 & 0.8103 & 0.8138 & 0.6724 & 0.9483 \\
 & SALMONN & 0.6724 & 0.8021 & 0.6724 & 0.6330 & 0.3448 & 1.0000 \\
\bottomrule
\end{tabular}
\caption{Zero-shot performance on Audio Plausibility (Generative Models). $\uparrow$ indicates higher is better.}
\label{tab:plausibility_generative_models_results}
\end{table*}

\begin{table*}[t]
\centering
\small
\begin{tabular}{llccccccc}
\toprule
\textbf{Dataset} & \textbf{ALM} & \textbf{Acc} $\uparrow$ & \textbf{P} $\uparrow$ & \textbf{R} $\uparrow$ & \textbf{F1} $\uparrow$ & \textbf{E-Acc} $\uparrow$ & \textbf{N-Acc} $\uparrow$ & \textbf{C-Acc} $\uparrow$ \\
\midrule
\multirow{8}{*}{Afri-200} & AudioFlamingo2 & 0.3333 & 0.1111 & 0.3333 & 0.1667 & 0.0000 & 1.0000 & 0.0000 \\
 & AudioFlamingo3 & 0.6367 & 0.7398 & 0.6367 & 0.5466 & 0.9800 & 0.0400 & 0.8900 \\

 & GAMA & 0.2967 & 0.1739 & 0.2967 & 0.1936 & 0.8400 & 0.0500 & 0.0000 \\
 & Kimi & 0.6333 & 0.7499 & 0.6333 & 0.5383 & 0.8700 & 0.0700 & 0.9600 \\
 & Qwen2.5 Omni & 0.6800 & 0.7076 & 0.6800 & 0.6731 & 0.8600 & 0.3400 & 0.8400 \\
 & Qwen2 Audio 7B & 0.7133 & 0.7283 & 0.7133 & \textbf{0.7123} & 0.7400 & 0.5300 & 0.8700 \\
 & SALMONN & 0.3967 & 0.4520 & 0.3967 & 0.2814 & 0.0000 & 1.0000 & 0.1900 \\
\addlinespace[0.5em]
\multirow{8}{*}{Medical} & AudioFlamingo2 & 0.3057 & 0.1024 & 0.3278 & 0.1561 & 0.0000 & 0.9833 & 0.0000 \\
 & AudioFlamingo3 & 0.6218 & 0.5512 & 0.5954 & 0.5160 & 0.9863 & 0.0167 & 0.7833 \\
 & GAMA & 0.3782 & 0.2621 & 0.3591 & 0.3030 & 0.6438 & 0.4333 & 0.0000 \\
 & Kimi & 0.6500 & 0.6726 & 0.6500 & \textbf{0.5995} & 0.8250 & 0.2000 & 0.9250 \\
 & Qwen2.5 Omni & 0.5492 & 0.6414 & 0.5335 & 0.5274 & 0.7671 & 0.5500 & 0.2833 \\
 & Qwen2 Audio 7B & 0.5521 & 0.5802 & 0.5546 & 0.5534 & 0.5139 & 0.6833 & 0.4667 \\
 & SALMONN & 0.3679 & 0.6042 & 0.3737 & 0.2960 & 0.2877 & 0.8000 & 0.0333 \\
\bottomrule
\end{tabular}
\caption{Zero-shot performance on Audio Entailment (Generative Models). $\uparrow$ indicates higher is better.}
\label{tab:nli_generative_models_results}
\end{table*}

\begin{table*}[t]
\centering
\small
\begin{tabular}{llcccccc}
\toprule
\textbf{Dataset} & \textbf{ALM} & \textbf{Acc} $\uparrow$ & \textbf{P} $\uparrow$ & \textbf{R} $\uparrow$ & \textbf{F1} $\uparrow$ & \textbf{Acc\_C} $\uparrow$ & \textbf{Acc\_I} $\uparrow$ \\
\midrule
\multirow{8}{*}{Afri-200} & AudioFlamingo2 & 0.6019 & 0.6493 & 0.6019 & 0.5677 & 0.3204 & 0.8835 \\
 & AudioFlamingo3 & 0.5000 & 0.2500 & 0.5000 & 0.3333 & 0.0000 & 1.0000 \\
 & GAMA & 0.2573 & 0.7802 & 0.2573 & 0.2819 & 0.4951 & 0.0194 \\
 & Kimi & 0.8301 & 0.8545 & 0.8301 & 0.8271 & 0.6990 & 0.9612 \\
 & Qwen2.5 Omni & 0.7136 & 0.8179 & 0.7136 & 0.6880 & 1.0000 & 0.4272 \\
 & Qwen2 Audio 7B & 0.8447 & 0.8452 & 0.8447 & \textbf{0.8446} & 0.8641 & 0.8252 \\
 & SALMONN & 0.7379 & 0.7393 & 0.7379 & 0.7375 & 0.7767 & 0.6990 \\
\addlinespace[0.5em]
\multirow{8}{*}{Afri-Gen} & AudioFlamingo2 & 0.8793 & 0.9022 & 0.8793 & 0.8815 & 0.7759 & 0.9828 \\
 & AudioFlamingo3 & 0.5000 & 0.2500 & 0.5000 & 0.3333 & 0.0000 & 1.0000 \\
 & GAMA & 0.2759 & 0.8684 & 0.2759 & 0.3562 & 0.4828 & 0.0690 \\
 & Kimi & 0.8190 & 0.8418 & 0.8190 & 0.8159 & 0.6897 & 0.9483 \\
 & Qwen2.5 Omni & 0.9138 & 0.9265 & 0.9138 & \textbf{0.9131} & 1.0000 & 0.8276 \\
 & Qwen2 Audio7B & 0.8276 & 0.8625 & 0.8276 & 0.8233 & 0.6724 & 0.9828 \\
 & SALMONN & 0.9052 & 0.9053 & 0.9052 & 0.9052 & 0.8966 & 0.9138 \\
\addlinespace[0.5em]
\multirow{8}{*}{Medical} & AudioFlamingo2 & 0.8125 & 0.8175 & 0.8125 & \textbf{0.8118} & 0.7500 & 0.8750 \\
 & AudioFlamingo3 & 0.5000 & 0.2500 & 0.5000 & 0.3333 & 0.0000 & 1.0000 \\
 & GAMA & 0.3500 & 0.7429 & 0.3500 & 0.4210 & 0.5500 & 0.1500 \\
 & Kimi & 0.7375 & 0.8279 & 0.7375 & 0.7181 & 0.4750 & 1.0000 \\
 & Qwen2.5Omni & 0.7250 & 0.8226 & 0.7250 & 0.7025 & 1.0000 & 0.4500 \\
 & Qwen2 Audio  & 0.7625 & 0.7640 & 0.7625 & 0.7622 & 0.7250 & 0.8000 \\
 & SALMONN & 0.7750 & 0.7757 & 0.7750 & 0.7749 & 0.8000 & 0.7500 \\
\bottomrule
\end{tabular}
\caption{Zero-shot performance on Audio Consistency (Generative Models). $\uparrow$ indicates higher is better.}
\label{tab:consistency_generative_models_results}
\end{table*}

\begin{table*}[t]
\centering
\small
\begin{tabular}{llccccccc}
\toprule
\textbf{Dataset} & \textbf{ALM} & \textbf{Acc} $\uparrow$ & \textbf{P} $\uparrow$ & \textbf{R} $\uparrow$ & \textbf{F1} $\uparrow$ & \textbf{E-Acc} $\uparrow$ & \textbf{N-Acc} $\uparrow$ & \textbf{C-Acc} $\uparrow$ \\
\midrule
\multirow{3}{*}{Afri-200} & LAION-CLAP & 0.3333 & 0.3333 & 0.3333 & 0.3325 & 0.3000 & 0.3200 & 0.3800 \\
 & MSCLAP\_23 & 0.3200 & 0.2127 & 0.3200 & 0.2321 & 0.2000 & 0.0000 & 0.7600 \\
 & MSCLAP\_22 & 0.3500 & 0.3448 & 0.3500 & 0.3444 & 0.2200 & 0.4200 & 0.4100 \\
\addlinespace[0.5em]
\multirow{3}{*}{Medical} & LAION-CLAP & 0.3782 & 0.3868 & 0.3868 & 0.3725 & 0.2603 & 0.5500 & 0.3500 \\
 & MSCLAP\_23 & 0.3005 & 0.2019 & 0.3064 & 0.2246 & 0.2192 & 0.0000 & 0.7000 \\
 & MSCLAP\_22 & 0.3834 & 0.3988 & 0.3607 & 0.3286 & 0.6986 & 0.1167 & 0.2667 \\
\bottomrule
\end{tabular}
\caption{Zero-shot performance on audio entailment (Contrastive Models). $\uparrow$ indicates higher is better.}
\label{tab:nli_constructive_models_results}
\end{table*}

\section{Discussion}
\label{sec:discussion}

The empirical results prompt a number of takeaways about the current state of audio–semantic reasoning.  First, even the best-performing models exhibit systematic over-entailment: hypotheses that are plausible but not entailed by the audio are frequently marked as true.  This suggests that models rely heavily on linguistic priors and commonsense knowledge rather than strictly grounding their predictions in the acoustic evidence.  Second, domain shift remains a challenge.  Performance on the Medical corpus, which contains spontaneous clinical dialogues with colloquial phrasing, lags behind performance on the read-speech AfriSpeech-Gen set.  Models that excel on one domain may falter on another, indicating that domain-specific fine-tuning or broader training corpora may be necessary.

Accent variation further complicates semantic reasoning.  The accent-drift experiments reveal that models often infer different meanings from semantically equivalent utterances when those utterances are spoken with different accents.  Such bias underscores the risk of deploying audio systems in multilingual, multicultural settings without careful calibration.  The accent restraint task, which measures hallucination on low-semantic-content utterances, shows that most models readily invent content when little is present.  Among generative models, Qwen2-based variants strike a better balance between sensitivity and restraint, but there is ample room for improvement.

Finally, the stark contrast between generative and contrastive models highlights the limits of embedding-only approaches for semantic reasoning.  Although contrastive models perform well on retrieval-style tasks, they struggle to make fine-grained entailment judgments.  Future work might explore hybrid architectures or training objectives that explicitly align representations with reasoning labels.  We hope that the benchmark and analyses presented here will spur more nuanced evaluation and lead to audio language models that reason as carefully as they speak.

\section{Conclusion}
We have introduced a comprehensive benchmark for evaluating semantic reasoning in audio language models across diverse domains, tasks, and accents.  By combining large-scale hypothesis generation with careful human verification, our datasets isolate distinct reasoning phenomena including entailment, plausibility, consistency, accent-conditioned drift and semantic restraint that challenge current models beyond transcription quality.  Our results show that next-token prediction models, particularly Qwen2-based variants and AudioFlamingo, substantially outperform contrastive baselines on most reasoning tasks, yet still exhibit significant over-entailment and susceptibility to commonsense bias.  The benchmark also surfaces performance gaps under domain shift and accent variation, underscoring the need for robustness to linguistic diversity.  We hope that these resources and analyses will catalyze the development of audio language models that ground their inferences in acoustic evidence and handle the richness of spoken language with care.

\section*{Limitations}

This work has several limitations.  First, although we curated diverse datasets across multiple African accents and domains, the benchmark still reflects a finite selection of languages, topics and speakers; performance may not generalize to other dialects, spontaneous discourse genres or low-resource languages outside our coverage.  Second, while our human verification protocol mitigates hallucinated hypotheses, annotator judgments and edits remain subjective; some subtle distinctions could be missed or misclassified.  Third, the reliance on large language models for hypothesis generation and label mapping introduces their own biases and may not capture all error modes.  Finally, our evaluation focuses on zero-shot inference; fine-tuning or prompting strategies may yield different patterns of success and failure.  Future work should expand the dataset to additional languages and domains, incorporate multiple annotation phases for quality control, and explore training-time interventions to reduce over-entailment and accent sensitivity.


\bibliography{custom, references}

\appendix

\section{Dataset Details}
\label{sec:appendix_datasets}

We utilize a diverse collection of Pan-African speech and text datasets to evaluate semantic reasoning across clinical and general domains. Table~\ref{tab:datasets-detailed} provides a consolidated overview of the dataset statistics and accessibility.

\begin{table}[ht]
\centering
\small
\begin{tabular}{lccc}
\hline
\textbf{Dataset} & \textbf{Primary Domain} \\
\hline
AfriSpeech-200\  & Clinical/General ASR \\
AfriSpeech-Dialog\ &  Conversational ASR/SLU \\
Afri-Names\ &  Named Entity/Commands \\
Afrispeech-Medical\ &  Clinical Dialogues \\
\hline
\end{tabular}
\caption{Summary of datasets. These resources support evaluation across entailment, consistency, plausibility, and accent-conditioned reasoning tasks.}
\label{tab:datasets-detailed}
\end{table}

\paragraph{AfriSpeech-200} 
This corpus contains 67,577 audio–transcript pairs from 2,463 unique speakers across 13 African countries. It serves as the benchmark for reasoning tasks that require high-fidelity transcription of diverse African accents in both medical and everyday contexts.

\paragraph{AfriSpeech-Dialog} 
This dataset consists of simulated spontaneous conversations. Unlike read-speech corpora, it captures the nuances of natural disfluencies and conversational flow, specifically curated to test the robustness of spoken language understanding (SLU) models.

\paragraph{Afri-Names} 
A specialized read-speech dataset containing 6,307 samples. It focuses on high-precision segments—specifically numbers, African named entities, and voice commands—spanning 12 accents across 4 countries. It is used primarily to evaluate accent-conditioned reasoning and NER.

\paragraph{Med-Convo-Nig} 
Sourced from real-world Nigerian doctor–patient tele-consultations, this dataset provides high-stakes clinical dialogues. It is critical for evaluating model performance on local linguistic variations in healthcare-specific semantic tasks, such as clinical consistency and entailment.

\section{Human Verification Protocol}
\label{sec:appendix_humanverification}

All candidate hypotheses produced by the language model were audited by a team of three trained annotators with backgrounds in linguistics and speech technology.  The annotators followed a detailed guideline document that emphasised:

\begin{itemize}
    \item Listening to the entire audio premise, including pauses and prosodic cues, before reading the hypothesis.
    \item Classifying each hypothesis as supported, contradicted or unsupported based on the audio alone, without using external world knowledge.
    \item Editing hypotheses that contained hallucinated details, ambiguous wording, or reliance on unstated background information.  Edits were constrained to preserve the intended semantic relationship and remain faithful to the original audio.
    \item Flagging any problematic audio segments (e.g., very low signal-to-noise ratio) for exclusion from the benchmark.
\end{itemize}

Each hypothesis was seen by two annotators; disagreements were resolved through discussion with a third annotator acting as arbiter.  Random spot checks were performed by one of the authors to ensure consistency across the entire corpus.  This human verification step proved especially valuable for accented speech and unfamiliar proper names, where automatic generation alone often led to subtle semantic drift.

\section{Evaluation Metrics}
\label{app:evaluation_metrics}

We evaluate audio--semantic reasoning performance using a combination of
classification, calibration, and robustness-oriented metrics, following prior
work on audio entailment and audio--language model evaluation. Metrics are
computed at the hypothesis level and aggregated per model and dataset.

\subsection{Primary Classification Metrics}

For multi-class and binary decision tasks, we report standard classification
metrics derived from the confusion matrix.

\paragraph{Accuracy (ACC).}
Accuracy measures the proportion of correctly classified instances:
\begin{equation}
\mathrm{ACC} = \frac{TP + TN}{TP + TN + FP + FN}
\end{equation}
where $TP$, $TN$, $FP$, and $FN$ denote true positives, true negatives, false
positives, and false negatives, respectively.

Accuracy is reported for all tasks but may obscure class imbalance effects,
especially in entailment-style settings.

\paragraph{Precision (P).}
Precision measures the proportion of predicted positives that are correct:
\begin{equation}
\mathrm{P} = \frac{TP}{TP + FP}
\end{equation}

\paragraph{Recall (R).}
Recall measures the proportion of true positives that are correctly identified:
\begin{equation}
\mathrm{R} = \frac{TP}{TP + FN}
\end{equation}

\paragraph{F1 Score (F1).}
The F1 score is the harmonic mean of precision and recall:
\begin{equation}
\mathrm{F1} = \frac{2 \cdot \mathrm{P} \cdot \mathrm{R}}{\mathrm{P} + \mathrm{R}}
\end{equation}

For multi-class tasks (e.g., entailment, neutral, contradiction), we report
macro-averaged precision, recall, and F1 unless otherwise stated.

\subsection{Entailment-Aware Accuracy Metrics}

Following prior audio entailment benchmarks, we additionally report
label-specific accuracies to capture asymmetric error behavior.

\paragraph{Entailment Accuracy (EACC).}
\begin{equation}
\mathrm{EACC} = \frac{\text{\# correctly predicted entailment instances}}
{\text{\# entailment instances}}
\end{equation}

\paragraph{Neutral Accuracy (NACC).}
\begin{equation}
\mathrm{NACC} = \frac{\text{\# correctly predicted neutral instances}}
{\text{\# neutral instances}}
\end{equation}

\paragraph{Contradiction Accuracy (CACC).}
\begin{equation}
\mathrm{CACC} = \frac{\text{\# correctly predicted contradiction }}
{\text{\# contradiction}}
\end{equation}

These metrics are especially informative for diagnosing model bias toward
entailment or over-rejection behavior.

\subsection{Task Applicability of Metrics}

The applicability of each metric varies by task:

\begin{itemize}
    \item \textbf{Audio Entailment.}  
    We report ACC, P, R, F1, along with EACC, NACC, and CACC.

    \item \textbf{Audio--Text Semantic Consistency (Binary).}  
    We report ACC, P, R, and F1. Label-specific accuracies reduce to positive and
    negative class accuracy.

    \item \textbf{Semantic Plausibility Judgment.}  
    We report ACC and F1. Precision and recall are emphasized for the
    \emph{implausible} class to measure hallucination susceptibility.

\end{itemize}

\subsubsection{Robustness and Over-Inference Metrics (AfriNames)}

For AfriNames-style datasets, which lack rich semantic content, standard NLI
metrics are insufficient. Instead, we focus on restraint and robustness.

\paragraph{Semantic Restraint Accuracy (SRA).}
Semantic restraint accuracy measures the proportion of cases in which the model
correctly refrains from asserting unsupported semantic content:
\begin{equation}
\mathrm{SRA} = \frac{\text{\# correct neutral or abstain predictions}}
{\text{\# total AfriNames instances}}
\end{equation}

This multi-metric evaluation provides a comprehensive and task-aligned assessment
of audio--language models across diverse spoken domains.

\subsection{Handling Multiple Hypotheses per Label}
\label{app:multiple_hypotheses}

For several datasets and tasks, including Audio--Text Semantic Consistency and Spoken Natural Language Inference, we generate multiple hypotheses per semantic label (e.g., two consistent and two inconsistent hypotheses per audio instance). This design choice reflects the inherent variability in how a single semantic relation can be linguistically expressed, while maintaining a fixed underlying label.

\paragraph{Evaluation Strategy.}
We adopt a unified evaluation strategy in which all hypotheses sharing the same semantic label are treated equivalently. Specifically, each hypothesis--audio pair is evaluated independently by the model, and the resulting predictions are aggregated at the label level. For binary tasks such as semantic consistency, hypotheses labeled as \textit{consistent} are treated as positive instances, while those labeled as \textit{inconsistent} are treated as negative instances. For ternary tasks such as entailment, hypotheses are grouped into \textit{entails}, \textit{neutral}, and \textit{contradicts} categories.

This approach allows us to preserve hypothesis-level diversity without requiring separate inference runs or task-specific evaluation pipelines.

\paragraph{Implications for Metrics and Tables.}
All reported metrics (Accuracy, Precision, Recall, F1, and class-conditional accuracies) are computed over the full set of hypothesis--audio pairs. As a result, performance tables (e.g., Table~4) summarize model behavior at the \emph{task level}, rather than at the individual hypothesis variant level. Each table corresponds to a single task (e.g., entailment or consistency), aggregating over all hypothesis realizations within that task.

This design ensures that evaluation remains comparable across datasets and models, while avoiding fragmentation into multiple near-duplicate tables.

\paragraph{Duplicate Hypothesis and Correlation Considerations.}
Although multiple hypotheses are generated from the same transcript, they are not duplicates in the lexical sense. Each hypothesis differs in phrasing, abstraction level, or semantic emphasis, even when sharing the same label. Consequently, models must generalize across paraphrastic variation rather than exploit surface-level repetition.

To mitigate concerns about correlation between hypotheses derived from the same audio, we emphasize that:
(i) hypotheses are evaluated independently,
(ii) metrics are reported in aggregate rather than per-audio averages, and
(iii) our primary goal is to assess semantic robustness under linguistic variability rather than estimate per-instance uncertainty.

\section{Different LLMs and their Generated Hypotheses}
\label{sec:appendix_qualitative}

To illustrate qualitative differences among next-token prediction models, we include a set of manually curated examples in Table~\ref{tab:qualitative_examples}.  Each block shows the original transcript followed by hypotheses generated by Llama~3.1~8B, Mistral~Large~3 and Qwen~2~7B, annotated according to the corresponding semantic reasoning task.  These examples highlight how different model architectures navigate plausibility, entailment, consistency, accent-induced drift and restraint.

\begin{table}[h]
\centering
\small
\begin{tabularx}{\columnwidth}{lp{0.5cm}l}
\toprule
\textbf{Model} & \textbf{Model Size} & \textbf{LLM Text Decoder} \\
\midrule
Qwen2.5-Omni & 7B & Qwen2.5-7B-Instruct  \\
Qwen2-Audio & 7B & Qwen2-7B-Instruct  \\
SALMONN & 13B & Vicuna-13B \\
Kimi-Audio & 12.5B & Kimi-LLM-v1  \\
GAMA & 7B & Vicuna-7B (v1.5)  \\
AudioFlamingo2 & 3B & Qwen2.5-3B  \\
AudioFlamingo3 & 7B & Qwen2.5-7B  \\
MSCLAP 2022 & - & HTSAT / RoBERTa  \\
MSCLAP 2023 & - & HTSAT / RoBERTa  \\
LAION-CLAP & - & Swin / RoBERTa  \\
\bottomrule
\end{tabularx}
\caption{Architectural specifications of evaluated Audio Language Models.}
\label{tab:model_specs}
\end{table}

\begin{table*}[t]
\centering
\small
\begin{tabular}{p{2.5cm} p{2.5cm} p{10cm}}
\toprule
\textbf{Dataset / Task} & \textbf{Model} & \textbf{Transcript and Generated Hypotheses} \\
\midrule

\multicolumn{3}{l}{\textbf{AfriNames — Accent Drift (Over-Inference)}} \\

\multicolumn{3}{p{15cm}}{\textit{Transcript:}
“In the environmental assessment conducted over 34 months, researchers identified 24 endangered species and proposed a …”
} \\

 & Llama~3.1~8B &
 \textbf{Accent-Invariant:} The assessment identified multiple endangered species requiring protection. \newline
 \textbf{Accent-Sensitive Lure:} The researchers are likely affiliated with a prestigious Western university. \\

 & Mistral~Large~3 &
 \textbf{Accent-Invariant:} The study revealed significant threats to biodiversity. \newline
 \textbf{Accent-Sensitive Lure:} The findings were shaped by international conservation agendas. \\

 & Qwen~2~7B &
 \textbf{Accent-Invariant:} Several species were found to be at risk of extinction. \newline
 \textbf{Accent-Sensitive Lure:} The research was conducted in an urbanized region. \\

\midrule
\multicolumn{3}{l}{\textbf{AfriNames — Accent Restraint (Semantic Minimality)}} \\

\multicolumn{3}{p{15cm}}{\textit{Transcript:}
“The 1921 census projects a population surge to 488 million, indicating a 4\% growth rate influenced by immigration.”
} \\

 & Llama~3.1~8B &
 \textbf{Supported:} The population is projected to increase. \newline
 \textbf{Unsupported:} The census results will directly influence national economic policy. \\

 & Mistral~Large~3 &
 \textbf{Supported:} Immigration contributes to population growth. \newline
 \textbf{Unsupported:} The census was designed to justify housing reforms. \\

 & Qwen~2~7B &
 \textbf{Supported:} A population surge is anticipated. \newline
 \textbf{Unsupported:} The growth rate is unusually high for a developed country. \\

\midrule
\multicolumn{3}{l}{\textbf{AfriSpeech-200 — Semantic Consistency}} \\

\multicolumn{3}{p{15cm}}{\textit{Transcript:}
“This study builds on her previous work examining what qualities and skills make for stronger middle-grade science teachers.”
} \\

 & Llama~3.1~8B &
 \textbf{Consistent:} The study focuses on identifying characteristics of effective science teachers. \newline
 \textbf{Inconsistent:} The study explores the impact of dance training on teachers. \\

 & Mistral~Large~3 &
 \textbf{Consistent:} The research examines skills linked to strong science instruction. \newline
 \textbf{Inconsistent:} The work analyzes athletic performance in schools. \\

 & Qwen~2~7B &
 \textbf{Consistent:} The paper investigates qualities that improve science teaching. \newline
 \textbf{Inconsistent:} The research evaluates music education programs. \\

\midrule
\multicolumn{3}{l}{\textbf{AfriSpeech-200 — Audio Entailment}} \\

\multicolumn{3}{p{15cm}}{\textit{Transcript:}
“Snow arriving from the north later tomorrow, moving south through Thursday morning may lead to some travel disruption.”
} \\

 & Llama~3.1~8B &
 \textbf{Entailment:} Weather conditions may disrupt travel over the next two days. \newline
 \textbf{Contradiction:} Travel will not be affected by weather conditions. \\

 & Mistral~Large~3 &
 \textbf{Entailment:} Snowfall could interfere with transportation schedules. \newline
 \textbf{Contradiction:} Clear weather will ensure uninterrupted travel. \\

 & Qwen~2~7B &
 \textbf{Entailment:} Adverse weather may impact travel plans. \newline
 \textbf{Contradiction:} No travel disruptions are expected. \\

\midrule
\multicolumn{3}{l}{\textbf{AfriSpeech-200 — Semantic Plausibility}} \\

\multicolumn{3}{p{15cm}}{\textit{Transcript:}
“Not everyone can run a business effectively, so it is important that younger shoemakers are trained to be recruited by big companies.”
} \\

 & Llama~3.1~8B &
 \textbf{Plausible:} Training increases employment opportunities for younger shoemakers. \newline
 \textbf{Implausible:} Large companies avoid hiring trained shoemakers. \\

 & Mistral~Large~3 &
 \textbf{Plausible:} Professional training prepares shoemakers for industry roles. \newline
 \textbf{Implausible:} Shoemaking skills are irrelevant to corporate recruitment. \\

 & Qwen~2~7B &
 \textbf{Plausible:} Skill development improves recruitment prospects. \newline
 \textbf{Implausible:} Shoemaking is no longer valued by employers. \\

\bottomrule
\end{tabular}
\caption{Qualitative examples illustrating how different large language models generate task-specific hypotheses from the same spoken input. Each block presents the original transcript followed by model-generated hypotheses corresponding to the semantic reasoning task.}
\label{tab:qualitative_examples}
\end{table*}
This table specifies the underlying LLM decoder and the total parameter count (where known) for the 10 models.

\section{Hyperparameters}
This table captures your specific sampling settings and the retry logic. Note that for CLAP-based models, generation parameters are not applicable, as they rely on similarity. All Experiments were ran on a single Nvidia Blackwell B200 and 80GB of RAM

\begin{table*}[h]
\centering
\small
\begin{tabular}{lccccc}
\toprule
\textbf{Model} & \textbf{Max New Tokens} & \textbf{Temp} & \textbf{Top-K} & \textbf{Top-P} & \textbf{Audio Temp/K} \\
\midrule
Qwen2.5-Omni & 512 & 0.0 & 5 & - & - / - \\
Qwen2-Audio & 512 & 0.0 & 5 & - & - / - \\
SALMONN & 512 & - & - & - & - / - \\
Kimi-Audio & 512 & 0.0 & 5 & - & 0.8 / 10 \\
GAMA & 400 & 0.1 & - & 0.95 & - / - \\
AudioFlamingo2 & 512 & 0.0 & 5 & - & - / - \\
AudioFlamingo3 & 512 & 0.0 & 5 & - & - / - \\
LAION-CLAP & - & - & - & - & - / - \\
MSCLAP 2023 & - & - & - & - & - / - \\
MSCLAP 2022 & - & - & - & - & - / - \\
\bottomrule
\end{tabular}
\caption{Inference hyperparameters. Generative models employ a greedy retry strategy (top\_k=1) if initial sampling fails. CLAP models use similarity-based retrieval.}
\label{tab:hyperparameters}
\end{table*}

\begin{table*}[t]
\centering
\small
\begin{tabular}{llcccccc}
\toprule
\textbf{Dataset} & \textbf{ALM} & \textbf{Acc} $\uparrow$ & \textbf{P} $\uparrow$ & \textbf{R} $\uparrow$ & \textbf{F1} $\uparrow$ & \textbf{Acc\_P} $\uparrow$ & \textbf{Acc\_I} $\uparrow$ \\
\midrule
\multirow{3}{*}{Afri-200} & LAION-CLAP & 0.4976 & 0.4968 & 0.4970 & 0.4906 & 0.6117 & 0.3824 \\
 & MSCLAP\_23 & 0.5122 & 0.5124 & 0.5124 & \textbf{0.5116} & 0.4757 & 0.5490 \\
 & MSCLAP\_22 & 0.4976 & 0.4987 & 0.4997 & 0.3841 & 0.0680 & 0.9314 \\
\addlinespace[0.5em]
\multirow{3}{*}{Afri-Gen} & LAION-CLAP & 0.5603 & 0.5626 & 0.5603 & \textbf{0.5564} & 0.6552 & 0.4655 \\
 & MSCLAP\_23 & 0.5172 & 0.5176 & 0.5172 & 0.5149 & 0.5862 & 0.4483 \\
 & MSCLAP\_22 & 0.5431 & 0.5550 & 0.5431 & 0.5169 & 0.3103 & 0.7759 \\
\addlinespace[0.5em]
\multirow{3}{*}{Medical} & LAION-CLAP & 0.5875 & 0.6068 & 0.5875 & \textbf{0.5680} & 0.8000 & 0.3750 \\
 & MSCLAP\_23 & 0.4000 & 0.3990 & 0.4000 & 0.3985 & 0.4500 & 0.3500 \\
 & MSCLAP\_22 & 0.5375 & 0.5419 & 0.5375 & 0.5250 & 0.3750 & 0.7000 \\
\bottomrule
\end{tabular}
\caption{Zero-shot performance on audio Plausibility (Contrastive Models). $\uparrow$ indicates higher is better.}
\label{tab:plausibility_constructive_models_results}
\end{table*}

\begin{table*}[t]
\centering
\small
\begin{tabular}{llcccccc}
\toprule
\textbf{Dataset} & \textbf{ALM} & \textbf{Acc} $\uparrow$ & \textbf{P} $\uparrow$ & \textbf{R} $\uparrow$ & \textbf{F1} $\uparrow$ & \textbf{Acc\_C} $\uparrow$ & \textbf{Acc\_I} $\uparrow$ \\
\midrule
\multirow{3}{*}{Afri-200} & LAION-CLAP & 0.5049 & 0.5050 & 0.5049 & \textbf{0.5018} & 0.4272 & 0.5825 \\
 & MSCLAP\_23 & 0.4660 & 0.4467 & 0.4660 & 0.4128 & 0.7670 & 0.1650 \\
 & MSCLAP\_22 & 0.5340 & 0.5514 & 0.5340 & 0.4908 & 0.2427 & 0.8252 \\
\addlinespace[0.5em]
\multirow{3}{*}{Afri-Gen} & LAION-CLAP & 0.5603 & 0.5618 & 0.5603 & \textbf{0.5577} & 0.6379 & 0.4828 \\
 & MSCLAP\_23 & 0.4483 & 0.4483 & 0.4483 & 0.4483 & 0.4483 & 0.4483 \\
 & MSCLAP\_22 & 0.5000 & 0.5000 & 0.5000 & 0.3967 & 0.0862 & 0.9138 \\
\addlinespace[0.5em]
\multirow{3}{*}{Medical} & LAION-CLAP & 0.5750 & 0.5758 & 0.5750 & 0.5739 & 0.5250 & 0.6250 \\
 & MSCLAP & 0.6250 & 0.6374 & 0.6250 & \textbf{0.6164} & 0.7750 & 0.4750 \\
 & MSCLAP\_22 & 0.5000 & 0.5000 & 0.5000 & 0.4972 & 0.5750 & 0.4250 \\
\bottomrule
\end{tabular}
\caption{Zero-shot performance on Audio Consistency (Contrastive Models). $\uparrow$ indicates higher is better.}
\label{tab:consistency_constructive_models_results}
\end{table*}

\begin{table*}[t]
\centering
\small
\begin{tabular*}{\textwidth}{@{\extracolsep{\fill}} lll ccccccc}
\toprule
\textbf{Model} & \textbf{Diff.} & \textbf{Acc} $\uparrow$ & \textbf{P} $\uparrow$ & \textbf{R} $\uparrow$ & \textbf{F1} $\uparrow$ & \textbf{E-Acc} $\uparrow$ & \textbf{N-Acc} $\uparrow$ & \textbf{C-Acc} $\uparrow$ \\
\midrule
Cascade Whisper + Llama-3.1-8B & easy & 0.565 & 0.624 & 0.561 & 0.542 & 0.913 & 0.542 & 0.227 \\
 & medium & 0.594 & 0.629 & 0.602 & 0.571 & 0.913 & 0.727 & 0.167 \\
 & hard & 0.522 & 0.597 & 0.522 & 0.495 & 0.913 & 0.478 & 0.174 \\
\addlinespace[0.5em]
Cascade Whisper + Mistral-7B & easy & 0.449 & 0.712 & 0.460 & 0.412 & 0.304 & 0.167 & 0.909 \\
 & medium & 0.435 & 0.708 & 0.423 & 0.361 & 0.261 & 0.091 & 0.917 \\
 & hard & 0.522 & 0.772 & 0.522 & 0.485 & 0.435 & 0.174 & 0.957 \\
\addlinespace[0.5em]
Cascade Whisper + Qwen2.5-7B & easy & 0.812 & 0.834 & 0.807 & 0.806 & 0.913 & 0.917 & 0.591 \\
 & medium & 0.681 & 0.694 & 0.687 & 0.679 & 0.826 & 0.818 & 0.417 \\
 & hard & 0.725 & 0.748 & 0.725 & 0.726 & 0.870 & 0.696 & 0.609 \\
\addlinespace[0.5em]
Cascade Granite + Llama-3.1-8B & easy & 0.478 & 0.582 & 0.475 & 0.436 & 0.913 & 0.375 & 0.136 \\
 & medium & 0.536 & 0.610 & 0.544 & 0.492 & 0.957 & 0.591 & 0.083 \\
 & hard & 0.420 & 0.541 & 0.420 & 0.341 & 0.957 & 0.261 & 0.043 \\
\addlinespace[0.5em]
Cascade Granite + Mistral-7B & easy & 0.449 & 0.789 & 0.462 & 0.391 & 0.261 & 0.125 & 1.000 \\
 & medium & 0.464 & 0.752 & 0.453 & 0.382 & 0.087 & 0.273 & 1.000 \\
 & hard & 0.362 & 0.668 & 0.362 & 0.246 & 0.087 & 0.043 & 0.957 \\
\addlinespace[0.5em]
Cascade Granite + Qwen2.5-7B & easy & 0.725 & 0.733 & 0.723 & 0.726 & 0.696 & 0.792 & 0.682 \\
 & medium & 0.667 & 0.666 & 0.670 & 0.667 & 0.696 & 0.773 & 0.542 \\
 & hard & 0.754 & 0.753 & 0.754 & 0.753 & 0.739 & 0.870 & 0.652 \\
\bottomrule
\end{tabular*}
\caption{Detailed performance of cascaded ASR+LLM systems on Audio Entailment (Interview NLI) across difficulty levels.}
\label{tab:cascade_nli_appendix}
\end{table*}

\begin{table*}
   \centering
\small
\begin{tabular}{llcc}
\toprule
\textbf{Model} & \textbf{Bias Rate} $\downarrow$ & \textbf{Accept Rate} $\uparrow$ \\
\midrule
AudioFlamingo2 & 96.75\% & 97.75\% \\
AudioFlamingo3 & 81.50\% & \textbf{98.25\%} \\
GAMA & 58.75\% & 68.25\% \\
Kimi & 55.25\% & 84.50\% \\
Qwen2.5Omni & 41.50\% & 95.50\% \\
Qwen2AudioInstruct & 33.25\% & 88.75\% \\
SALMONN & 1.50\% & 3.50\% \\
\bottomrule
\end{tabular}
\caption{AfriNames accent-drift evaluation across models. $\downarrow$ indicates lower is better; $\uparrow$ indicates higher is better.}
\label{tab:afrinames_accent_drift} 
\end{table*}

\begin{table*}[t]
\centering
\small
\begin{tabular}{llccc}
\toprule
\textbf{Model} & \textbf{HLU Rate} $\downarrow$ & \textbf{SPRT Rate} $\uparrow$ & \textbf{SRA} $\uparrow$ \\
\midrule

AudioFlamingo3 & 88.00\% & 100.00\% & 66.00\% \\
GAMA & 96.83\% & 98.50\% & \textbf{72.62\%} \\
Kimi & 17.17\% & 80.50\% & 12.88\% \\
Qwen2.5Omni & 2.33\% & 62.50\% & 1.75\% \\
Qwen2AudioInstruct & 39.17\% & 97.50\% & 29.38\% \\
SALMONN & 27.17\% & 35.50\% & 20.38\% \\
LAION-CLAP & 60.17\% & 68.00\% & 45.12\% \\
\bottomrule
\end{tabular}
\caption{AfriNames restraint evaluation across models. The SRA ($\uparrow$) metric balances the model's ability to identify supported content against its tendency toward hallucination ($\downarrow$). HLU stands for Hallucination and SPRT stands for support} 
\label{tab:afrinames_restraint}
\end{table*}

\section{Prompts}
\label{app:prompts}
In section we show all the prompts used by the Large Language Model and prompts used by the Audio Language model. 

\begin{table*}[t]
    \centering
    \begin{tcolorbox}[
        colback=white, 
        colframe=black!80, 
        sharp corners, 
        boxrule=0.8pt, 
        width=\textwidth,
        title=\textbf{Prompt 1: Semantic Restraint / Over-Inference Generation}
    ]
        \small
        You are a careful evaluator of spoken language meaning. You are given a verbatim transcript of a very short spoken audio recording. This audio contains LIMITED semantic content. 
        
        \vspace{0.4em} \hrule \vspace{0.5em}

        \textbf{TASK: Generate the following:}
        \begin{enumerate}[leftmargin=*, nosep]
            \item \textbf{SUPPORTED hypotheses (2 items)} – Statements clearly supported; minimal and cautious.
            \item \textbf{UNSUPPORTED hypotheses (6 items)} – Statements that sound reasonable but are NOT supported; designed to tempt over-inference.
        \end{enumerate}

        \vspace{0.5em}
        \textbf{CONSTRAINTS:}
        \begin{itemize}[leftmargin=*, nosep]
            \item Do NOT quote/paraphrase transcript or describe what the speaker ``says.''
            \item Do NOT add roles, topics, locations, intent, or domains unless explicitly supported.
            \item Avoid explicit negation terms (``no,'' ``not,'' ``never'').
        \end{itemize}

        \vspace{0.5em} \hrule \vspace{0.5em}

        \textbf{GOAL:} Unsupported hypotheses should test for: Invented topics, intent, social roles, domains, and situational context.

        \vspace{0.5em}
        \textbf{OUTPUT FORMAT (JSON):} \\
        \texttt{\{ "supported": ["...", "..."], "unsupported": ["...", "...", "...", "...", "...", "..."] \}}

        \vspace{0.5em}
        \textbf{TRANSCRIPT:} \texttt{\{\{TRANSCRIPT\}\}}
    \end{tcolorbox}
    \caption{Prompt used to generate diagnostic items for evaluating model hallucination and over-inference.}
    \label{prompt:semantic_restraint}
\end{table*}

\begin{table*}[t]
    \centering
    \begin{tcolorbox}[
        colback=white, 
        colframe=black!80, 
        sharp corners, 
        boxrule=0.8pt, 
        width=\textwidth,
        title=\textbf{Prompt 2: Accent-Conditioned Semantic Drift Generation}
    ]
        \small
        You are evaluating semantic stability in spoken language understanding. \textbf{IMPORTANT:} This task tests whether models infer DIFFERENT meanings from similar short utterances spoken with different accents or pronunciations.
        
        \vspace{0.4em} \hrule \vspace{0.5em}

        \textbf{TASK: Generate the following:}
        \begin{enumerate}[leftmargin=*, nosep]
            \item \textbf{ACCENT-INVARIANT hypotheses (4 items)} – Truth should NOT depend on accent or speaker identity; rely only on semantic content.
            \item \textbf{ACCENT-SENSITIVE LURE hypotheses (4 items)} – Statements models might incorrectly infer due to accent, pronunciation, or training bias.
        \end{enumerate}

        \vspace{0.5em}
        \textbf{CONSTRAINTS:}
        \begin{itemize}[leftmargin=*, nosep]
            \item Do NOT assume speaker nationality, ethnicity, gender, age, or location.
            \item Do NOT quote/paraphrase. Each hypothesis must be one complete sentence.
            \item Avoid explicit negation terms.
        \end{itemize}

        \vspace{0.5em} \hrule \vspace{0.5em}

        \textbf{GOAL:} Lures should test for: Assigned social roles, geographic/cultural backgrounds, and injected domain meanings.

        \vspace{0.5em}
        \textbf{OUTPUT FORMAT (JSON):} \\
        \texttt{\{ "accent\_invariant": [...], "accent\_sensitive\_lures": [...] \}}

        \vspace{0.5em}
        \textbf{TRANSCRIPT:} \texttt{\{\{TRANSCRIPT\}\}}
    \end{tcolorbox}
    \caption{Prompt used to generate diagnostic items for evaluating fairness and stability under accent variation.}
    \label{prompt:accent_drift}
\end{table*}

\begin{figure*}[t]
    \centering
    \begin{tcolorbox}[
        colback=white, colframe=black!80, sharp corners, boxrule=0.8pt, width=\textwidth,
        title=\textbf{Semantic Restraint / Over-Inference Prompt}
    ]
        \small
        \textbf{Purpose:} Generate hypotheses to test whether models correctly withhold inference when audio content is minimal. \\
        \textbf{Methodological Value:} Measures hallucination/over-inference rates and provides a clean signal for semantic restraint.
        \vspace{0.4em} \hrule \vspace{0.5em}

        \textbf{System Prompt:}
        You are a careful evaluator of spoken language meaning. You are given a verbatim transcript of a very short spoken audio recording. This audio contains LIMITED semantic content. Your task is to generate hypotheses that test whether a model can correctly WITHHOLD inference.

        \textbf{Task:}
        Generate the following:
        \begin{itemize}[leftmargin=*, nosep]
            \item \textbf{SUPPORTED hypotheses (2x):} Clearly supported by the audio; minimal and cautious.
            \item \textbf{UNSUPPORTED hypotheses (2x):} Likely to tempt a model to over-infer; not supported by this specific audio.
        \end{itemize}

        \textbf{Constraints:}
        \begin{itemize}[leftmargin=*, nosep]
            \item Do not quote/paraphrase; no meta-language (``the speaker says'').
            \item Do not add roles, topics, intent, or domains unless explicitly supported.
            \item \textbf{Prohibited:} Explicit negation terms (``no'', ``not'', ``never'', ``without'').
        \end{itemize}

        \vspace{0.5em} \hrule \vspace{0.5em}
        \textbf{Output Format (JSON):}
        \texttt{\{ "supported": ["...", "..."], "unsupported": ["...", "..."] \}}
    \end{tcolorbox}
    \label{prompt:semantic-restraint-over-inference}
\end{figure*}

\begin{figure*}[t]
    \centering
    \begin{tcolorbox}[
        colback=white, colframe=black!80, sharp corners, boxrule=0.8pt, width=\textwidth,
        title=\textbf{Accent-Conditioned Semantic Drift Prompt}
    ]
        \small
        \textbf{Purpose:} Test if models infer different meanings from similar utterances spoken with different accents. \\
        \textbf{Methodological Value:} Measures fairness gaps and semantic drift triggered by non-semantic audio features.
        \vspace{0.4em} \hrule \vspace{0.5em}

        \textbf{System Prompt:}
        You are evaluating semantic stability. This task tests whether models infer DIFFERENT meanings from similar short utterances spoken with different accents or pronunciations.

        \textbf{Task:}
        \begin{itemize}[leftmargin=*, nosep]
            \item \textbf{ACCENT-INVARIANT (4x):} Truth value depends only on semantics, not speaker identity.
            \item \textbf{ACCENT-SENSITIVE LURES (4x):} Incorrect inferences likely caused by accent or training bias.
        \end{itemize}

        \textbf{Constraints:}
        \begin{itemize}[leftmargin=*, nosep]
            \item Do not assume nationality, ethnicity, gender, or location.
            \item Do not use explicit negation terms; no quoting the transcript.
        \end{itemize}

        \vspace{0.5em} \hrule \vspace{0.5em}
        \textbf{Output Format (JSON):}
        \verb|{ "accent_invariant": [...], "accent_sensitive_lures": [...] }|
    \end{tcolorbox}
    \label{prompt:accent-conditioned-semantic-drift}
\end{figure*}

\begin{figure*}[t]
    \centering
    \begin{tcolorbox}[
        colback=white, colframe=black!80, sharp corners, boxrule=0.8pt, width=\textwidth,
        title=\textbf{Audio Entailment}
    ]
        \small
        \textbf{System Prompt:}
        You are a helpful assistant with expert knowledge in spoken language understanding and parliamentary discourse. Generate hypotheses for a Spoken Natural Language Inference (NLI) task based on the provided institutional transcript.

        \textbf{Instructions:}
        Generate three hypotheses for each:
        \begin{itemize}[leftmargin=*, nosep]
            \item \textbf{Entailment:} Definitely true given the meaning of the spoken audio.
            \item \textbf{Neutral:} Might be true but cannot be determined from audio alone.
            \item \textbf{Contradiction:} Definitely false given the meaning of the audio.
        \end{itemize}

        \textbf{Constraints:}
        \begin{itemize}[leftmargin=*, nosep]
            \item Reflect semantic inference, not surface wording; do not quote the transcript.
            \item \textbf{Prohibited:} Explicit negation terms (``no'', ``not'', ``never'').
        \end{itemize}

        \vspace{0.5em} \hrule \vspace{0.5em}
        \textbf{Output Format (JSON):}
        \texttt{\{ "entailment": [...], "neutral": [...], "contradiction": [...] \}}
    \end{tcolorbox}
    \label{prompt:audio-entailment}
\end{figure*}

\begin{figure*}[t]
    \centering
    \begin{tcolorbox}[
        colback=white, colframe=black!80, sharp corners, boxrule=0.8pt, width=\textwidth,
        title=\textbf{Audio--Text Semantic Consistency Prompt}
    ]
        \small
        \textbf{System Prompt:}
        You are assisting in the construction of an audio–text semantic consistency dataset using parliamentary audio transcripts.

        \textbf{Task:}
        Generate four text statements:
        \begin{itemize}[leftmargin=*, nosep]
            \item \textbf{Consistent (2x):} Semantically aligned with the audio.
            \item \textbf{Inconsistent (2x):} Semantically conflicting with the audio.
        \end{itemize}

        \textbf{Constraints:}
        \begin{itemize}[leftmargin=*, nosep]
            \item Concern institutional roles, procedures, or formal discourse.
            \item Do not quote or restate the transcript; do not use explicit negation words.
        \end{itemize}

        \vspace{0.5em} \hrule \vspace{0.5em}
        \textbf{Output Format (JSON):}
        \texttt{\{ "consistent": ["...", "..."], "inconsistent": ["...", "..."] \}}
    \end{tcolorbox}
    \label{prompt:audio-text-semantic-consistency}
\end{figure*}

\begin{figure*}[t]
    \centering
    \begin{tcolorbox}[
        colback=white, colframe=black!80, sharp corners, boxrule=0.8pt, width=\textwidth,
        title=\textbf{Audio Semantic Plausibility Prompt}
    ]
        \small
        \textbf{System Prompt:}
        You are given a transcript of a spoken medical audio recording.

        \textbf{Task:}
        Generate four statements:
        \begin{itemize}[leftmargin=*, nosep]
            \item \textbf{Plausible (2x):} Align with the medical context and clinical commonsense.
            \item \textbf{Implausible (2x):} Unlikely given the audio context and medical norms.
        \end{itemize}

        \textbf{Constraints:}
        \begin{itemize}[leftmargin=*, nosep]
            \item Plausibility judgments may rely on medical commonsense.
            \item Do not quote the transcript; do not use explicit negation.
        \end{itemize}

        \vspace{0.5em} \hrule \vspace{0.5em}
        \textbf{Output Format (JSON):}
        \texttt{\{ "plausible": ["...", "..."], "implausible": ["...", "..."] \}}
    \end{tcolorbox}
    \label{prompt:audio-semantic-plausibility}
\end{figure*}


\begin{figure*}[t]
    \centering
    \begin{tcolorbox}[
        colback=white, colframe=black!80, sharp corners, boxrule=0.8pt, width=\textwidth,
        title=\textbf{ALM Prompt 1: Audio Entailment (Zero-Shot Classification)}
    ]
        \small
        \textbf{Context:} Evaluation of logical inference between a raw audio signal and a text hypothesis.
        \vspace{0.4em} \hrule \vspace{0.5em}

        \textbf{System Prompt:}
        You are an Audio Reasoning Agent. You are given an audio recording and a hypothesis. Determine the logical relationship between them.

        \textbf{Task:}
        Determine whether the hypothesis is:
        \begin{itemize}[leftmargin=*, nosep]
            \item \textbf{Entailed:} The hypothesis is a direct logical consequence of the audio.
            \item \textbf{Contradicted:} The hypothesis is logically impossible given the audio.
            \item \textbf{Neutral:} The hypothesis is plausible but neither supported nor refuted by the audio.
        \end{itemize}

        \textbf{Constraint:} Respond with exactly one label from the following set: \{\texttt{entailment}, \texttt{contradiction}, \texttt{neutral}\}.

        \vspace{0.5em} \hrule \vspace{0.5em}
        \textbf{Input Format:} 
        \verb!Audio: <audio_signal> | Hypothesis: {{HYPOTHESIS}}!
    \end{tcolorbox}
    \label{prompt:audio-entailment-zero-shot}
\end{figure*}

\begin{figure*}[t]
    \centering
    \begin{tcolorbox}[
        colback=white, colframe=black!80, sharp corners, boxrule=0.8pt, width=\textwidth,
        title=\textbf{ALM Prompt 2: Semantic Consistency (Binary Alignment)}
    ]
        \small
        \textbf{Context:} Determining the semantic alignment between auditory evidence and a descriptive statement.
        \vspace{0.4em} \hrule \vspace{0.5em}

        \textbf{System Prompt:}
        Given an audio recording and a statement, determine whether the statement is semantically consistent with the audio.

        \textbf{Constraint:} Respond with exactly one label: \{\texttt{consistent}, \texttt{inconsistent}\}. Do not provide explanations.

        \vspace{0.5em} \hrule \vspace{0.5em}
        \textbf{Input Format:} 
        \verb!Audio: <audio_signal> | Statement: {{STATEMENT}}!
    \end{tcolorbox}
    \label{prompt:alm-prompt-2}
\end{figure*}

\begin{figure*}[t]
    \centering
    \begin{tcolorbox}[
        colback=white, colframe=black!80, sharp corners, boxrule=0.8pt, width=\textwidth,
        title=\textbf{ALM Prompt 3: Semantic Plausibility (Contextual Reasoning)}
    ]
        \small
        \textbf{Context:} Assessing if a statement is likely/reasonable within the situational context of the audio.
        \vspace{0.4em} \hrule \vspace{0.5em}

        \textbf{System Prompt:}
        Given an audio recording and a statement, determine whether the statement is plausible given the audio context and general world knowledge.

        \textbf{Constraint:} Respond with exactly one label: \{\texttt{plausible}, \texttt{implausible}\}.

        \vspace{0.5em} \hrule \vspace{0.5em}
        \textbf{Input Format:} 
        \verb!Audio: <audio_signal> | Statement: {{STATEMENT}}!
    \end{tcolorbox}
    \label{prompt:alm-prompt-3}
\end{figure*}

\begin{figure*}[t]
    \centering
    \begin{tcolorbox}[
        colback=white, colframe=black!80, sharp corners, boxrule=0.8pt, width=\textwidth,
        title=\textbf{CLAP Evaluation Framework}
    ]
        \small
        \textbf{Methodology:} Unlike generative ALMs, CLAP evaluates semantic reasoning through similarity-based alignment between audio embeddings and templated text hypotheses. 
        \vspace{0.4em} \hrule \vspace{0.5em}

        \textbf{Templated Hypothesis Generation:}
        For each task, CLAP generates task-specific templates to transform labels into natural language assertions:
        \begin{itemize}[leftmargin=*, nosep]
            \item \textbf{Consistency:} \textit{``Given the audio, the following statement is consistent: \{hypothesis\}''}
            \item \textbf{Entailment:} \textit{``Given the audio, the following statement is true: \{hypothesis\}''}
            \item \textbf{Plausibility:} \textit{``Given the audio, the following statement is plausible: \{hypothesis\}''}
        \end{itemize}

        \vspace{0.5em}
        \textbf{Inference Mechanism:}
        The model computes cosine similarity scores $S(a, t_i)$ between the audio embedding ($a$) and each templated hypothesis embedding ($t_i$). The final prediction is determined by:
        \[ \text{Label} = \underset{i}{\arg\max}\ S(a, t_i) \]

        \vspace{0.5em} \hrule \vspace{0.5em}
        \textbf{Data Schema (Output JSON):}
        \begin{itemize}[leftmargin=*, nosep]
            \item \texttt{scores}: Vector of similarity scores for each label option (e.g., \textit{Consistent} vs. \textit{Inconsistent}).
            \item \texttt{best\_match\_idx}: Index of the hypothesis template with the highest similarity score.
            \item \texttt{best\_match\_hypothesis}: The full raw text of the highest-scoring templated statement.
            \item \texttt{best\_match\_label}: The categorical label extracted from the winning template.
            \item \texttt{pred}: The binary/multiclass prediction indicator for the sample.
        \end{itemize}
    \end{tcolorbox}
    \caption{The CLAP evaluation pipeline. This framework adapts contrastive audio-text models for reasoning tasks by wrapping hypotheses in semantic templates and selecting labels based on embedding similarity.}
    \label{fig:clap_appendix}
\end{figure*}

\begin{table*}[t]
    \centering
    \begin{tcolorbox}[
        colback=white, 
        colframe=black!80, 
        sharp corners, 
        boxrule=0.8pt, 
        width=\textwidth,
        title=\textbf{Prompt Variants for Semantic Consistency (v1--v3)}
    ]
        \small
        \textbf{(v1) Baseline Verification} \\
        Audio statement verification task. Does the following statement match what is said in the audio? \\
        \texttt{Statement: \{hypothesis\}} \\
        Answer with exactly one word -- either \texttt{CONSISTENT} or \texttt{INCONSISTENT}.

        \vspace{0.5em} \hrule \vspace{0.5em}

        \textbf{(v2) Detailed Instructions} \\
        Listen carefully to the audio recording. Then read the statement below. Your task: Decide if the statement MATCHES or CONTRADICTS what is said in the audio.
        \begin{itemize}[leftmargin=*, nosep]
            \item If the statement accurately reflects what is said in the audio, respond: \texttt{CONSISTENT}
            \item If the statement contradicts or misrepresents what is said in the audio, respond: \texttt{INCONSISTENT}
        \end{itemize}
        Think about what you heard in the audio before answering. \\
        \texttt{STATEMENT: \{hypothesis\}} \\
        Your answer (\texttt{CONSISTENT} or \texttt{INCONSISTENT}):

        \vspace{0.5em} \hrule \vspace{0.5em}

        \textbf{(v3) Formal Role Assignment} \\
        You are given an audio recording and a text statement. Determine whether the text is consistent with the meaning conveyed in the audio. Respond with one of the following labels only: \texttt{CONSISTENT} or \texttt{INCONSISTENT}. \\
        \texttt{STATEMENT: \{hypothesis\}}
    \end{tcolorbox}
    \caption{Three prompt variants used in the ablation study (Table~\ref{tab:prompt_ablation_af3}) to evaluate model sensitivity to prompt phrasing in semantic consistency tasks.}
    \label{app:prompt_variants_box}
\end{table*}

\begin{table*}[t]
\centering
\small
\setlength{\tabcolsep}{6pt}
\begin{tabular}{l l c c c c c c}
\toprule
\textbf{Dataset} & \textbf{Prompt Variant} & \textbf{Acc.} & \textbf{Prec.} & \textbf{Rec.} & \textbf{F1} & \textbf{Acc. (Con.)} & \textbf{Acc. (Inc.)} \\
\midrule
AfriSpeech-200 &
AudioFlamingo3 (v1) & 0.50 & 0.13 & 0.25 & 0.17 & 0.00 & 1.00 \\
AfriSpeech-200 &
AudioFlamingo3 (v2) & 0.53 & 0.52 & 0.36 & \textbf{0.28} & 0.07 & 1.00 \\
AfriSpeech-200 &
AudioFlamingo3 (v3) & 0.53 & 0.31 & 0.21 & 0.16 & 0.06 & 1.00 \\
\midrule
AfriSpeech-General &
AudioFlamingo3 (v1) & 0.50 & 0.17 & 0.33 & 0.22 & 0.00 & 1.00 \\
AfriSpeech-General &
AudioFlamingo3 (v2) & 0.54 & 0.24 & 0.16 & 0.14 & 0.09 & 1.00 \\
AfriSpeech-General &
AudioFlamingo3 (v3) & 0.76 & 0.42 & 0.38 & \textbf{0.37} & 0.52 & 1.00 \\
\midrule
Medical &
AudioFlamingo3 (v1) & 0.50 & 0.17 & 0.33 & 0.22 & 0.00 & 1.00 \\
Medical &
AudioFlamingo3 (v2) & 0.53 & 0.31 & 0.21 & 0.16 & 0.05 & 1.00 \\
Medical &
AudioFlamingo3 (v3) & 0.64 & 0.79 & 0.64 & \textbf{0.58} & 0.28 & 1.00 \\
\bottomrule
\end{tabular}
\caption{Prompt ablation for AudioFlamingo3 on the semantic consistency task. Best macro F1 per dataset is bolded.}
\label{tab:prompt_ablation_af3}
\end{table*}

\begin{table*}[t]
\centering
\small
\begin{tabular}{l cccc ccc}
\toprule
\textbf{ALM} & \textbf{Acc} $\uparrow$ & \textbf{P} & \textbf{R} & \textbf{F1} & \textbf{E-Acc} & \textbf{N-Acc} & \textbf{C-Acc} \\
\midrule
AudioFlamingo2 & 0.3109 & 0.1036 & 0.3333 & 0.1581 & 0.0000 & 1.0000 & 0.0000 \\
AudioFlamingo3 & 0.4663 & 0.5368 & 0.4802 & 0.4872 & 0.2740 & 0.4333 & 0.7333 \\
Kimi           & 0.3057 & 0.6168 & 0.3238 & 0.2490 & 0.0548 & 0.8167 & 0.1000 \\
Qwen2.5-Omni   & 0.4767 & 0.6568 & 0.4795 & 0.4673 & 0.4384 & 0.7667 & 0.2333 \\
Qwen2-Audio    & 0.3523 & 0.5423 & 0.3580 & 0.3036 & 0.2740 & 0.7500 & 0.0500 \\
\bottomrule
\end{tabular}
\caption{Zero-shot performance on Afrispeech Medical Entailment. All metrics are macro-averaged where applicable. $\uparrow$ indicates higher is better. The models struggle to caption and also reason all in the same prompt.}
\label{tab:nli_medical_results}
\end{table*}

\begin{figure*}[t]
    \centering
    \begin{tcolorbox}[
        colback=white, 
        colframe=black!80, 
        sharp corners, 
        boxrule=0.8pt, 
        width=\textwidth,
        title=\textbf{Prompt: Spoken Natural Language Inference}
    ]
        \small
        You are evaluating Spoken Natural Language Inference (Spoken NLI). \\
        You are given: (1) an audio recording (premise), (2) a text hypothesis.
        
        \vspace{0.4em} \hrule \vspace{0.5em}

        \textbf{Step 1 — AUDIO CAPTION (evidence only):} \\
        Write a short caption of what is explicitly said in the audio.
        \begin{itemize}[leftmargin=*, nosep]
            \item Use 1–2 sentences.
            \item Do NOT infer unstated details or add background knowledge.
            \item If audio is unclear, say ``unclear'' rather than guessing.
        \end{itemize}

        \vspace{0.5em}
        \textbf{Step 2 — REASONING (use caption as evidence):} \\
        Using ONLY the caption from Step 1 as evidence, determine whether the hypothesis is: \textit{ENTAILMENT}, \textit{CONTRADICTION}, or \textit{NEUTRAL}.
        
        \vspace{0.5em} \hrule \vspace{0.5em}

        \textbf{Output format:} \\
        \texttt{CAPTION: <your caption>} \\
        \texttt{LABEL: <ENTAILMENT|CONTRADICTION|NEUTRAL>} \\

        \vspace{0.5em}
        \textbf{Hypothesis:} ``<HYPOTHESIS\_TEXT>''
    \end{tcolorbox}
    \caption{Prompt structure for zero-shot Spoken Natural Language Inference using a caption-then-reasoning chain.}
    \label{fig:nli_prompt}
\end{figure*}

\end{document}